\documentclass[letterpaper, 10 pt, conference]{ieeeconf}  

\usepackage{amsmath}
\usepackage{hyperref}
\usepackage{graphicx}
\usepackage{caption}
\usepackage{graphicx}
\usepackage{tikz}
\usepackage{pgfplots}
\pgfplotsset{compat=1.10}
\usepackage{multirow}
\usepackage{subcaption}
\usepackage{siunitx}
\usepackage{algorithm}
\usepackage{algpseudocode}
\algnewcommand{\LineComment}[1]{\State \(\triangleright\) #1}
\usetikzlibrary{shapes.geometric, arrows, positioning}
\usepackage{cite}
\IEEEoverridecommandlockouts                              

\title{\LARGE \bf
Path Planning in Dynamic Environments using Generative RNNs and Monte Carlo Tree
Search}

\author{Stuart Eiffert, He Kong,  Navid Pirmarzdashti and Salah Sukkarieh
}

\begin{document}
\tikzstyle{rect} = [fill, rectangle, fill=white!20, text width=6em, text centered, minimum height=2em]
\tikzstyle{line} = [draw, -latex']

\maketitle
\thispagestyle{empty}
\pagestyle{empty}

\begin{abstract}
	
State of the art methods for robotic path planning in dynamic environments, such as crowds or traffic, rely on hand crafted motion models for agents. These models often do not reflect interactions of agents in real world scenarios. To overcome this limitation, this paper proposes an integrated path planning framework using generative Recurrent Neural Networks within a Monte Carlo Tree Search (MCTS). This approach uses a learnt model of social response to predict crowd dynamics during planning across the action space. This extends our recent work using generative RNNs to learn the relationship between planned robotic actions and the likely response of a crowd. We show that the proposed framework can considerably improve motion prediction accuracy during interactions, allowing more effective path planning. The performance of our method is compared in simulation with existing methods for collision avoidance in a crowd of pedestrians, demonstrating the ability to control future states of nearby individuals. We also conduct preliminary real world tests to validate the effectiveness of our method.
	







\end{abstract}

%

\section{Introduction}


Autonomous navigation among moving agents such as humans or self-driving
vehicles is becoming an increasingly important problem as we see more
applications of robotics in real world environments.
Predictive models of motion are required for planning around moving agents. It is crucial that these models reflect real world behaviours and capture how agents respond to the actions of a robot. This is needed to accurately understand how a robot influences it's environment, a critical requirement for safety, robustness, and efficiency in applications in dynamic environments, such as crowds of pedestrians and herds of livestock.

In this paper we propose an integrated path
planning framework using generative Recurrent Neural Networks (RNNs) and Monte Carlo Tree
Search (MCTS).
This builds upon recent work in motion prediction of crowds~\cite{Alahi2016,Vemula}. 
We use RNNs in an encoder-decoder architecture, where the input at each timestep is the 
current position of agent and the future position of the robot, which is known during 
training and sampled during planning.

This approach allows the encoding of observed sequences of agent positions and robot
actions, which can be used by the decoder stage to generate likely responses
of all agents to a robot's action. 
The generative model is used within the MCTS to simulate state transitions for sampled actions during a tree search of the robot's action space.
\begin{figure}
    \centering
	\includegraphics[width=8.5cm,height=5cm]{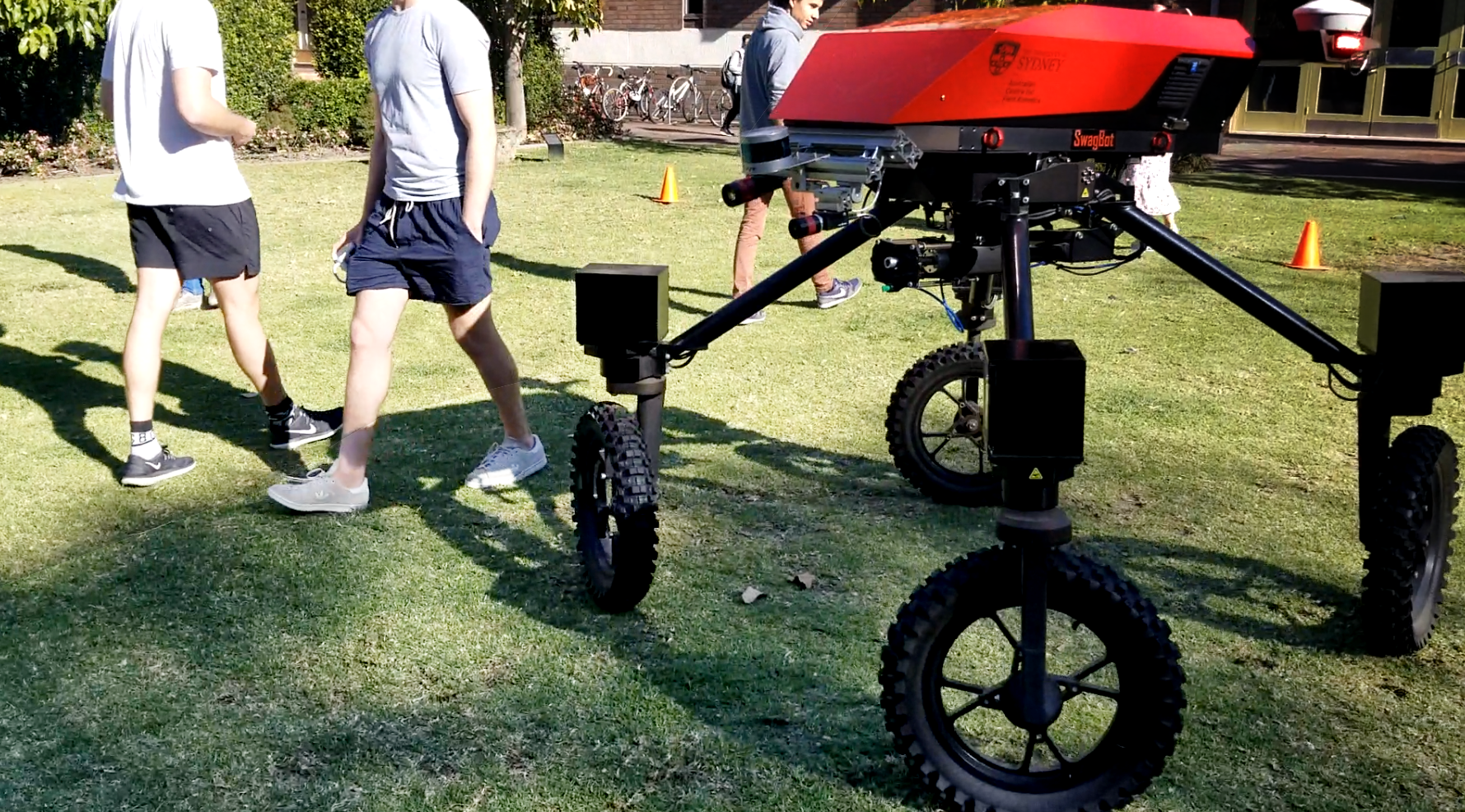}
	\caption{\textit{Real world testing of our MCTS-RNN dynamic path planner around moving pedestrians.}}
\end{figure}

We validate the proposed approach on three varied datasets, including 
interactions between pedestrians and a vehicle \cite{Yang2019}, 
livestock and a mobile robot \cite{JUnderwood2013}, and simulated
interactions of pedestrians generated using the optimal reciprocal collision avoidance (ORCA) model of motion
\cite{VanDenBerg2011}. 

The performance of our path planning method is compared
to existing approaches including a reactive potential field 
and deep reinforcement learning.
The results demonstrate that not only does our planning
algorithm perform comparably to state of the art methods for
collision avoidance, but more importantly, it is able to direct the future
states of nearby individuals using a motion model learnt from real world
data, allowing application to tasks such as planning paths that maneuver
nearby individuals, or herding of livestock towards a goal.
We have conducted preliminary tests of our approach on a real robotic platform around moving pedestrians.

\section{Related Work}
\subsection{Dynamic Path Planning}
There is extensive literature on path planning and motion prediction in dynamic environments, with a number of surveys available \cite{Rudenko2019,Hoy2015,S.BeckerR.HugW.Hbner}.
We briefly discuss the most relevant works here.

 A critical minimum requirement of dynamic path planning is avoiding collisions between the robot and it's surroundings, with the problem of collision avoidance receiving much attention in past years \cite{Fox1997,Blackmore2011,Fiorini}. However, it is widely acknowledged
that many path planning methods suffer from the \textquotedblleft freezing
robot\textquotedblright\ problem, i.e., when the surrounding environment
becomes increasingly complex, the robot loses the ability to find a path and
chooses to stop or acts erratically to avoid collisions \cite%
{Trautman2010}.

To overcome the above issue, traditional planning methods~have been 
extended to account for interactions between agents and predict their future
motion during robot navigation, incorporating the hand-crafted
Social-Force model (SFM)\cite{Helbing1995}. SFM has been applied to characterise cooperation and interactions 
between agents as interacting Gaussian processes \cite{Trautman2010}. 
Additionally, the velocity obstacle model of motion in \cite{Fiorini} has been generalized to ORCA\cite{VanDenBerg2011} for multiple agents. 
This approach has also been combined with Bayseian inference to adaptively learn parameters for individually observed agents in BRVO \cite{Kim2015}.

Reinforcement learning (RL) methods have also been applied to
collision avoidance in dynamic environments. These approaches
learn evaluation functions that can be applied to any observed state, and
navigation policies that aim to maximise the expected value along a path.
RL methods are often initialised using imitation learning in a supervised
approach.~For example, \cite{Chen2018} uses state-action-value pairs taken
from training episodes generated by ORCA, whilst \cite{Everett2018} uses
pairs generated by their previous RL policy CADRL \cite{Chen2016}, which was
in turn initialised using ORCA. A challenge in adopting RL for applications
such as path planning in dynamic environments is that the state transitions
are unknown. To alleviate the above issue, many approaches assume that the
state will evolve with constant velocity over short time periods ~\cite%
{Chen2017}, or that all agents follow the ORCA model~\cite{Chen2018}. There
has also been recent interest in using actor-critic algorithms to
learn the policy and value functions simultaneously, applying a combination
of the current learnt policy and other simple motion models to all agents
within the scene \cite{Everett2018}.

Nonetheless, most of these RL methods still make use of simple parametric
motion models in training, rather than using models learnt from real
world data, such as can be captured by state of the art trajectory
prediction models \cite{Rudenko2019}.

Recent developments in RNN-based trajectory prediction methods\cite{Alahi2016, Vemula} allow for improved prediction in crowded environments, outperforming parametric based methods such as SFM \cite{S.BeckerR.HugW.Hbner}. Inspired by these methods, in this paper we will demonstrate how a predictive model can be adapted for use in path planning, using an encoder-decoder RNN architecture with MCTS.




\subsection{MCTS for Solving Sequential Decision Making Problems}

Path planning in dynamic environments can be formulated as a sequential
decision making problem. By including all relevant agent dynamics in the current state, it can be framed as a Markov Decision Process (MDPs). This can be achieved by using a trajectory prediction model to encode the observed sequence in a hidden state of an RNN, which we detail further in \textit{Section III}. 

MCTS allows the solving of MDPs where we do not know the state transition matrix
but can simulate future states. This is achieved through random sampling of
the action space in a structured decision tree \cite{Browne2012}. This approach has previously been applied to robotic planning in tasks such as high level action selection for autonomous driving \cite{Paxton2017}, and  motion planning for active perception in unknown environments ~\cite{Arora2017, Best2016}.

The search tree consists of nodes representing each state, and edges representing actions.
Upper Confidence Bounds applied to trees (UCT)\cite{Kocsis2006} is often applied in MCTS to balance exploration and exploitation during node selection. In UCT, the value $V$ of a node is determined as:
\begin{equation}
V={\displaystyle{\frac{w_{i}}{n_{i}^c}}+c\sqrt{\frac{\ln n_{i}^p}{n_{i}^c}}}
\end{equation}
where, after the i-th move:
$w_{i}$ is the cumulative reward of the node considered;
$n_{i}^c$ and $n_{i}^p$ are number of times the child and parent nodes have been traversed; and
c is an exploration-exploitation balance parameter. MCTS allows for anytime optimisation and propagation of state uncertainty, which we utilise in this work for path planning, detailed further in \textit{Section III}.





\begin{figure*}[tbp]
\centering
\includegraphics[width=14.5cm,height=7.0cm]{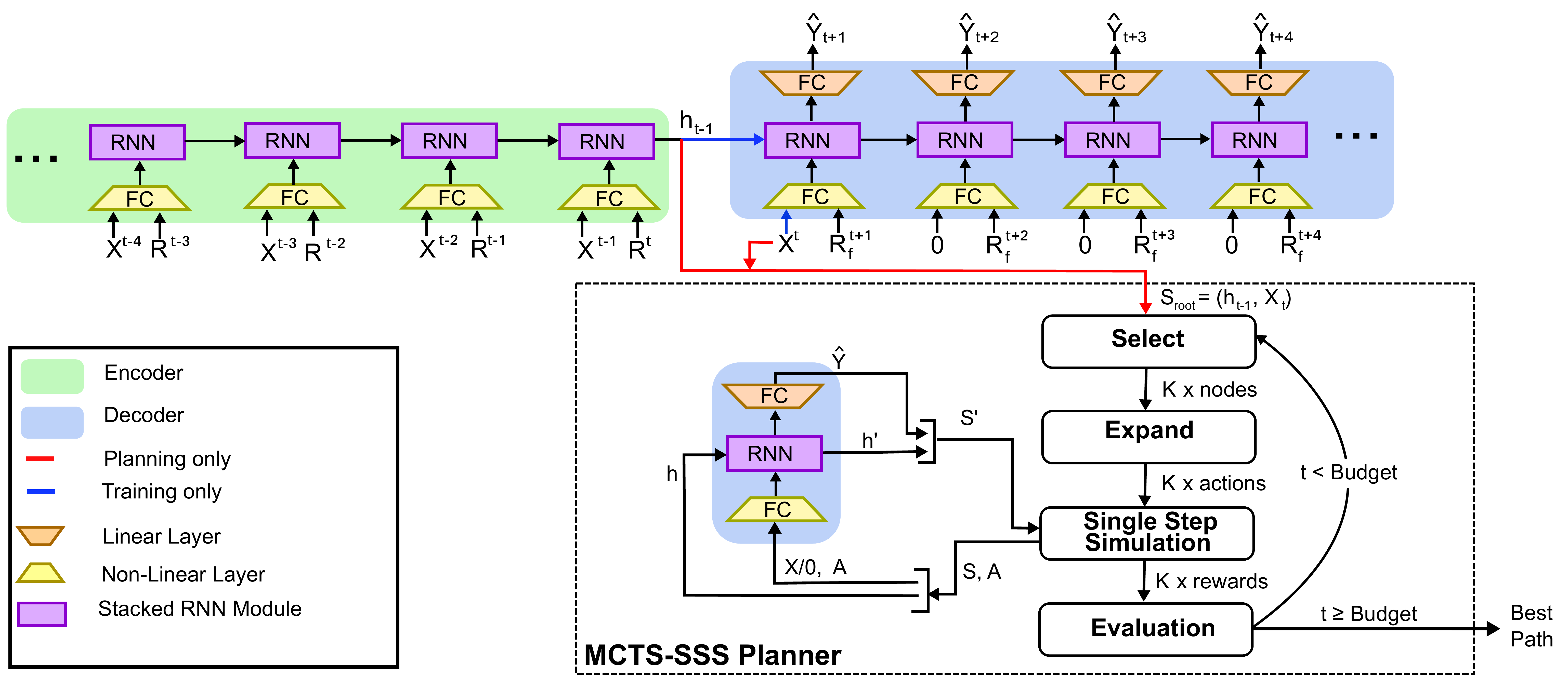}
\caption{\textit{System overview illustrating the use of a learnt model of
social response within a tree search based planner. After training (blue), the Encoder's final hidden state for a given observed sequence is used alongside the latest observation $X^t$ as the root state of the planner (red). The Decoder can then be used in a single step to simulate state transitions of the MDP for a given action \textit{A} and node state \textit{S}.}}
\end{figure*}

\section{Approach} 

\subsection{Overview}

Given observed trajectories $\textbf{X}$, and robot path $\textbf{R}$, for all timesteps in period $ t \leq T_{obs}$,  where $\textbf{X}^t= [\textbf{X}_1^t,\textbf{X}_2^t...,\textbf{X}_N^t]$ for $N$ non-controlled agents, we want to plan a path $\textbf{R}_p$ that optimises an objective function based both on the state of the robot and all other agents across a future time period $T_{obs} < t \leq T_{pred}$. The input trajectory for agent $i\in N$ is defined as it's position $X_i^t = [x_i^t,y_i^t]$ for each timestep $t$. $T_{obs}$ is the timestep of the latest observation, and $T_{pred}$ the future timestep to which we predict.

We achieve this by first training a sequence prediction model on $\textbf{X}$ and $\textbf{R}$, as well as the ground truth future positions of each agent $\textbf{Y}$ and known future positions of the robot $\textbf{R}_f$ for $T_{obs} < t \leq T_{pred}$.
We use this trained model in an adapted MCTS of the robot's action space across future timesteps, applied recursively within a receding horizon planner.


\textit{Fig. 2} illustrates the overall architecture of our approach, outlining the use of the  predictive model to first encode the observed trajectories for $t \leq T_{obs} -1$. The state of the root node of the MCTS $S_{root}$ is formed from  the final encoded state $h^{t-1}$ and the current observation $\textbf{X}^t$.
During the creation of the search tree, the predictive model is again used at each expanded node to predict the next state $S'$, given the state-action pair $(S,a)$ in the MCTS simulation step. Our integrated predictive planner is summarised in \textit{Alg. 1}.


\subsection{Learnt Model of Social Response}

%


The predictive model of response used in this work combines our prior work
ResponseRNN\cite{Eiffert2019} with the Recurrent Encoder-Decoder (RED)
model we used as a baseline comparison in the same work. The RED model is based on the Seq2Seq model in~\cite{S.BeckerR.HugW.Hbner}, which does not pool interactions between agents. 
Models that do not consider interactions between all agents in a scene have been shown to perform comparably to those which do, often outperforming them in terms of prediction accuracy even in crowded scenes \cite{Yang2019}. These simpler models also have inference speeds orders of magnitude less, which is critically important when conducting multiple predictions per timestep in a tree search approach. 

Our predictive model uses a robot's future action as an input, where the robot's position at timestep $t + \Delta t$ represents the action at time $t$. Thus, $\textbf{R}^{t+\Delta t}$ is used alongside $\textbf{X}^t$ as input to the predictive model, as shown in Fig. 2.

The use of a known future action as input has been shown to learn the
relationship between a robot's action and the likely next position of
nearby agents~\cite{Eiffert2019}. Experiment 1, in \textit{Section IV-A} details how the use of varying $\Delta t$ impacts prediction.

This model, learnt from real world interactions, can then be used to simulate state transitions in an approach similar to the use of parametric models in training deep RL.

\smallskip
\textbf{Model Structure:} The Encoder and Decoder have the same structure, comprised of a non linear embedding layer that takes in the input at each timestep, followed by two LSTM ~\cite{Hochreiter1997} layers. 
The inputs of the Encoder are made up of $\textbf{X}^t$ and $\textbf{R}^{t+\Delta t}$ for all $ t \leq T_{obs} -1 $.
The current observation at $t=T_{obs}$is used as the first input to the Decoder.

The Decoder takes the same size inputs as the Encoder, however, at all timesteps after the first Decoder input we feed the Decoder zeros in place of the agent positions. This is done for both training and inference. This \textit{zero-feed} approach has been shown to improve performance at inference time, when there are no known ground truth agent positions ~\cite{Zyner2018}. This is in comparison to other approaches that use a sample from the output of the prior step as input to the next step.
The non-linear embedding layer uses Rectified Linear unit (ReLu) activations and the same weights for both encoding and decoding steps. The outputs of the Decoder are passed through a linear layer that maps to  a bivariate Gaussian output for each agent's position at each predicted timestep.

\smallskip
\textbf{Training:} We use variable length encoding sequences between 8 and 20 timesteps. We decode for a fixed length of 8 timesteps, and compare the output of each Decoder step, $\hat {\textbf{Y}^t}$, to the ground truth positions of each agent $\textbf{Y}^t$. 
Training of the generative RNN is done so as to minimise the loss shown in \textit{Eq. 2},  which is the negative log-likelihood of $\textbf{Y}$  given $\hat {\textbf{Y}}$ , across all prediction timesteps.
\begin{equation}
Loss=-\sum_{t=T_{obs}+1}^{T_{pred}}\sum_{i}^{N} \log (P(x_{i}^{t},y_{i}^{t}|\hat{Y_{i}^{t}}))
\end{equation}
where $\hat{Y}_i^t=[\mu_{x_i}, \mu_{y_i}, \sigma_{x_i}, \sigma_{y_i}, \rho_i]^t$, 
for all agents $i\in N$, 
for all timesteps in prediction period $ T_{obs}+1 \leq t < T_{pred}$




\smallskip
\textbf{Inference:} 
Encoding is again performed on inputs across the period $ t \leq T_{obs} -1 $.
The final hidden state of the encoder, $h^{t-1}$, and the last observed agent positions $X^t$ at time $t=T_{obs}$ form the state of the root node of the tree search as $S_{root}=[h^{t-1},X^t]$. The Decoder is then used at each simulation step within the tree search to simulate $S'$, as detailed in the following section.

\subsection{Tree Search Planner}
MCTS is applied using UCT and adapted for parallel single step simulation (SSS).
\textit{Alg. 1} details the steps involved our adapted MCTS Planner.
We select the \textit{K} best nodes, according to the node value determined by \textit{Eq. 1}. 
We then expand each node, randomly choosing an action from its set of valid moves \textit{A}.
We next perform SSS in parallel for all chosen nodes, using the Decoder. This returns the predicted node's state $S'$, given the node's parent state $S=[h,X]$.
The cost of the node is then calculated according to the state evaluation function and propagated up through the tree. 
This process is repeated for a time budget, returning the best action from the root node.
\begin{algorithm}
	\caption{\strut Predictive Planner}\label{euclid}
	\begin{algorithmic}[1]
		\State A $\gets$ Actions \Comment{\textit{discretised action space}}
		\State B $\gets$ Budget \Comment{\textit{time in nsecs}}
		\State C $\gets$ CostFunction() \Comment{\textit{State Eval function}}
		\Function{MCTS-SSS(root, A, B, C)}{}
			\While{\texttt{time < B}} \Comment{\textit{planning budget}}
				\State K = Select(root) \Comment{\textit{Select K best nodes}}
				\State a = Expand(K, A) \Comment{\textit{Choose valid actions}}
				\LineComment{\textit{ parallel single step simulation}}
				\If {first iteration}
					\State $h',\hat{Y'}$ = RNN-Decoder($X^t$, a, h)
				\Else
					\State $h',\hat{Y'}$ = RNN-Decoder(0, a, h)
				\EndIf
				\State U = $\sqrt{det(cov(\hat{Y'}))}$ \Comment{\textit{uncertainty}}
				\State r = C($\hat{Y'}$,U ) \Comment{\textit{reward dependent on U}}
				\State Backup(K, r) \Comment{\textit{update node values}}
			\EndWhile\label{euclidendwhile}
			
			\Return{\texttt{root}}
		\EndFunction
		\While{\texttt{not at destination}}
			\State $X^{0:t}, R^{0:t-1}$ $\gets$ observe() 
			\LineComment{ \textit{ encode observed tracks}}
			\State $h^t,\hat{Y}^t$ = RNNEncoder($X^{0:t-1}, R^{0:t-1},h^0=0$)
			\State $S_{root} = (h^t,X^t)$ \Comment{\textit{create root node}}
			\LineComment{ \textit{ perform MCTS with SSS}}
			\State Tree = MCTS-SSS($S_{root},A,B,C$)
			\State $R_p$ = Tree.best\_plan() \Comment{\textit{Yield best current path}}
		\EndWhile\label{euclidendwhile}
	\end{algorithmic}
\end{algorithm}


\smallskip
\textbf{Parallel Single Step Simulation:}
We adapt the simulation stage of MCTS to terminate after a single step. This differs from normal implementation in which the episode plays out until a terminal state is reached, selecting random actions each iteration.
This is made possible by designing state evaluation functions capable of directly evaluating the
value of any observed state. Details of the state evaluation functions used are outlined in \textit{Section IV-B}.
This approach is similar to that used in game applications of MCTS ~\cite{Lorentz2016}, where it has been shown that limiting the steps of the simulation stage leads to improved performance when a node's value can be directly estimated.

This also allows parallelisation of the simulation stage, as all simulations now run for the same number of iterations and use the same Decoder model. We alter the selection stage to find the $K$ best
nodes to expand in the tree. This is implemented by updating the number of
traversals $n_{i}$ across each node in between selections, before we simulate. This results in a temporarily decreased value of the node as determined by the UCT method in \textit{Eq. 1}, and so a
decreased likelihood of selecting a node from the same branch. 

Simulation is then conducted across all $K$ nodes in parallel, where the selected actions $A$, and associated parent states $S$ are passed to the Decoder, returning the state of all expanded nodes $S'=(h',\hat{Y'})$, as shown in \textit{Alg. 1} line 10.

\smallskip
\textbf{Uncertainty:}
The use of a bivariate Gaussian distribution as output of the predictive
model provides a measure of uncertainty for each state estimation. 
We represent uncertainty $U$  using the square root of the determinant of the covariance matrix, $\Sigma$, of each output. This provides a measure of the volume of the ellipse defined by 1 standard deviation from the mean in all dimensions.
We denote $U=\sqrt{det(\Sigma )}$,  and then use $U$ in the state evaluation function to discount the reward
computed for each simulated future state.

\medskip
\section{Experiments and Results}
We conduct three separate experiments:
\begin{enumerate}
	\item The first aims to validate the use of a future robotic action as an input to a predictive model.
	\item The second compares the performance of our planner to traditional and state of the art approaches.
	\item Thirdly, we also test our path planner in a real world environment around moving pedestrians.
\end{enumerate}

\subsection{Predictive Model Validation}
We compare the average and final displacement errors (ADE and FDE) for our model trained with varying robot action lookaheads, $\Delta t \in$\{0,1,2,3,4,5\}. We also compare model accuracy when no robot action is included is the input.

\smallskip
\textbf{Datasets:}
This experiment has been repeated using three datasets:
\begin{itemize}
	\item Vehicle-Crowd Interaction (VCI) DUT~\cite{Yang2019}
	\item`A Robot Amongst the Herd' (ARATH)~\cite{JUnderwood2013} livestock-robot interactions
	\item Generated ORCA Trajectories
\end{itemize}
The two real world datasets have been chosen as they both contain interactions between a controlled vehicle and uncontrolled agents, and focus on agent-agent interactions, rather than agent-space interactions.
A description of the ARATH dataset and the method of data collection is detailed in~\cite{Eiffert2019}.
Both datasets have been preprocessed to have a frame rate of approximately 5Hz, with a mean of 0 and a standard deviation of 1 for all input dimensions.
The ORCA dataset has been included for use during planning experiments in the ORCA environment, and includes 10,000 randomly generated scenes of between 2 to 12 agents.
\begin{figure}
    \centering
	\includegraphics[width=9cm,height=12.5cm]{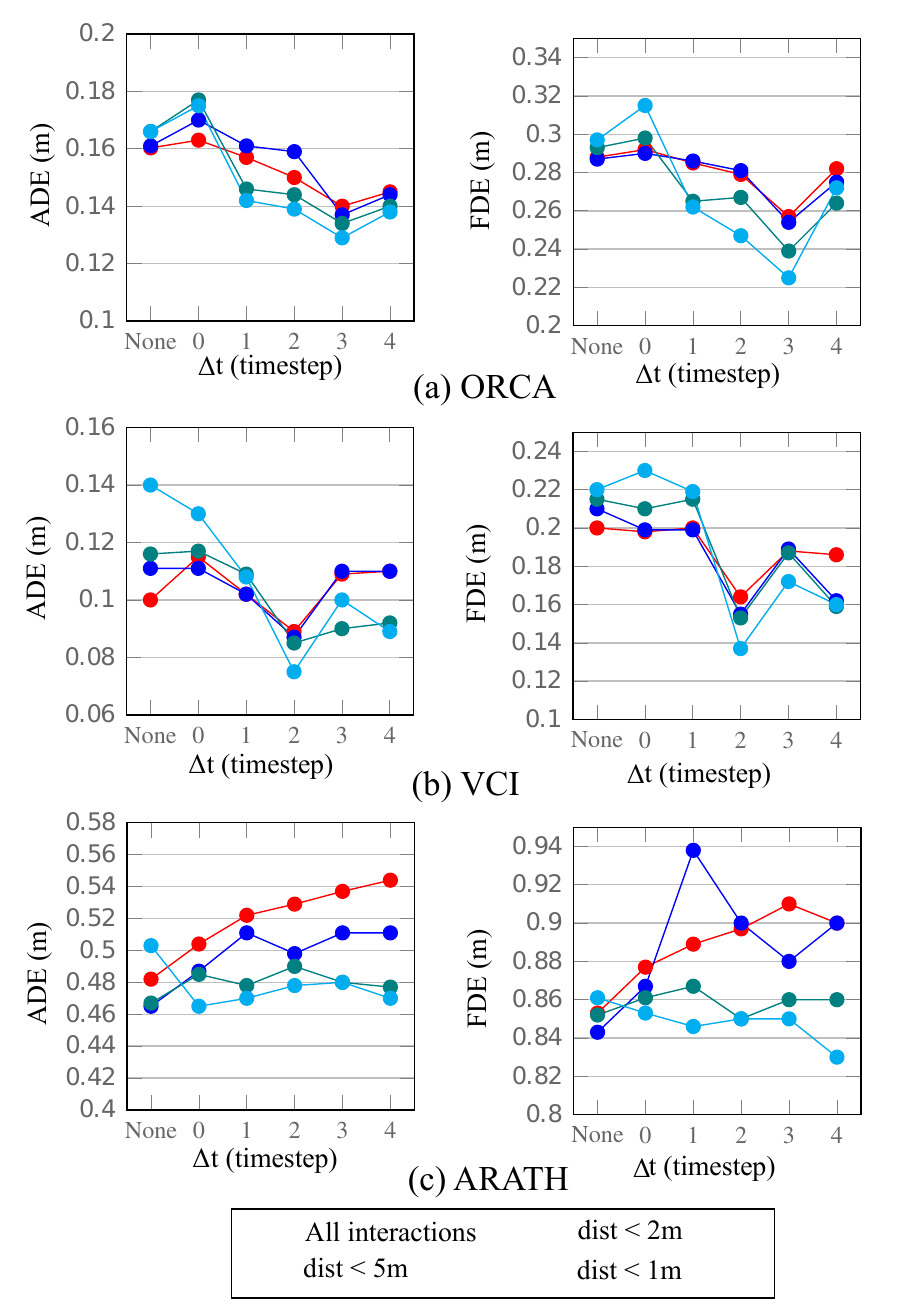}
	\caption{\textit{Comparison of prediction accuracy for varying robot lookaheads, $\Delta t$, and when no robot input is used ($\Delta t$=None). ADE and FDE are shown for all agents (red), and limited to agents near the robot (blue,cyan and teal). A clear correlation between $\Delta t$ and accuracy can be seen for ORCA and VCI, though not for ARATH except when distance limited.}}
\end{figure}

\smallskip
\textbf{Implementation:}
Each dataset has been split into 5 non-overlapping sets, of which 1 has been left for testing. We have used a 20\% validation split during training. The network is implemented in tensorflow with ADAM optimiser for 100 epochs on a single Titan-X GPU, taking approximately 1 hour to train. Inference time per decoder step is less than 0.1ms.

\smallskip
\textbf{Results:} \textit{Fig. 3} compares the prediction accuracy of our model trained using varying action lookaheads, $\Delta t$, as well as accuracy when no robot action is used.
For both pedestrian datasets, VCI and ORCA, there is a clear correlation between increased accuracy and $\Delta t >$ 0. This trend is more apparent when only considering the accuracy of agents within interaction distance thresholds of the robotic vehicle, shown for 5m, 2m and 1m thresholds. However, models trained on the ARATH livestock dataset do not display this same trend, only showing slight improvement of prediction when limited to distance thresholds of 2m and 1m.

This result confirms that using a planned robotic action as input to a predictive model of agent trajectory improves accuracy for close range interactions, accurately learning the relationship between a robot's planned action and the  response of an agent. This validates the use of such a model to simulate state transitions in a dynamic path planner. 

Whilst we include results for $\Delta t >$1 to display trends, we will only use $\Delta t$=1 in our planner comparisons.

\subsection{Planner Comparison}

We compare the performance of the following methods:
\begin{itemize}
	\item MCTS-RNN (Ours)
	\item LM-SARL (Local Map Self-Attention RL)~\cite{Chen2018}
	\item Reactive Potential Field (PF)
	\item MCTS-CV
\end{itemize}
MCTS-CV refers to our method with the predictive model is replaced by constant velocity motion for each agent.
We also compare performance of our method using a second state evaluation function, referred to as SEF2 in \textit{Table 1}.

\smallskip
\textbf{State Evaluation:}
We compare the following two state evaluation functions (SEF) for our planner:
\begin{flalign}
Cost &= (R^t - G)^2 + \sum_{i}^{N} \alpha {U_i^t}\\
Cost &= (R^t - G)^2 + \sum_{i}^{N} \alpha {U_i^t}(1+|\ddot{X}_i^t|),\\
     &\alpha= 
    \begin{cases}
        \frac{1}{X_i^t - R^t} ,& \text{if } X_i^t - R^t\leq d\\
        0,              & \text{otherwise}
    \end{cases} \notag
\end{flalign}    

where $R^t$ is the robot position, $G$ is the goal, $X^t_i$ is agent position and $U^t_i$ is prediction uncertainty for each agent $i\in N$. 
Both SEF1 (\textit{Eq. 3}) and SEF2 (\textit{Eq. 4}) apply a cost based on the distance of the robot to the goal, as well as the robot's distance to each agent, $\alpha$, which is scaled by the uncertainty of the agent's position and only considered when the robot is within a threshold distance $d$ of an agent. In SEF2 an additional term scales the cost based on the agent's acceleration when it is near the robot.
This term aims to limit the impact of the robot on the current velocity of each agent, and is referred to as \textit{Disturbance} in the results. SEF2 has been chosen to demonstrate the ability of our path planner to influence agents' future states without retraining the predictive model. For testing, we use $d=2m$.

\begin{figure*}[tbp]
\medskip
\includegraphics[width=17.5cm,height=8.0cm]{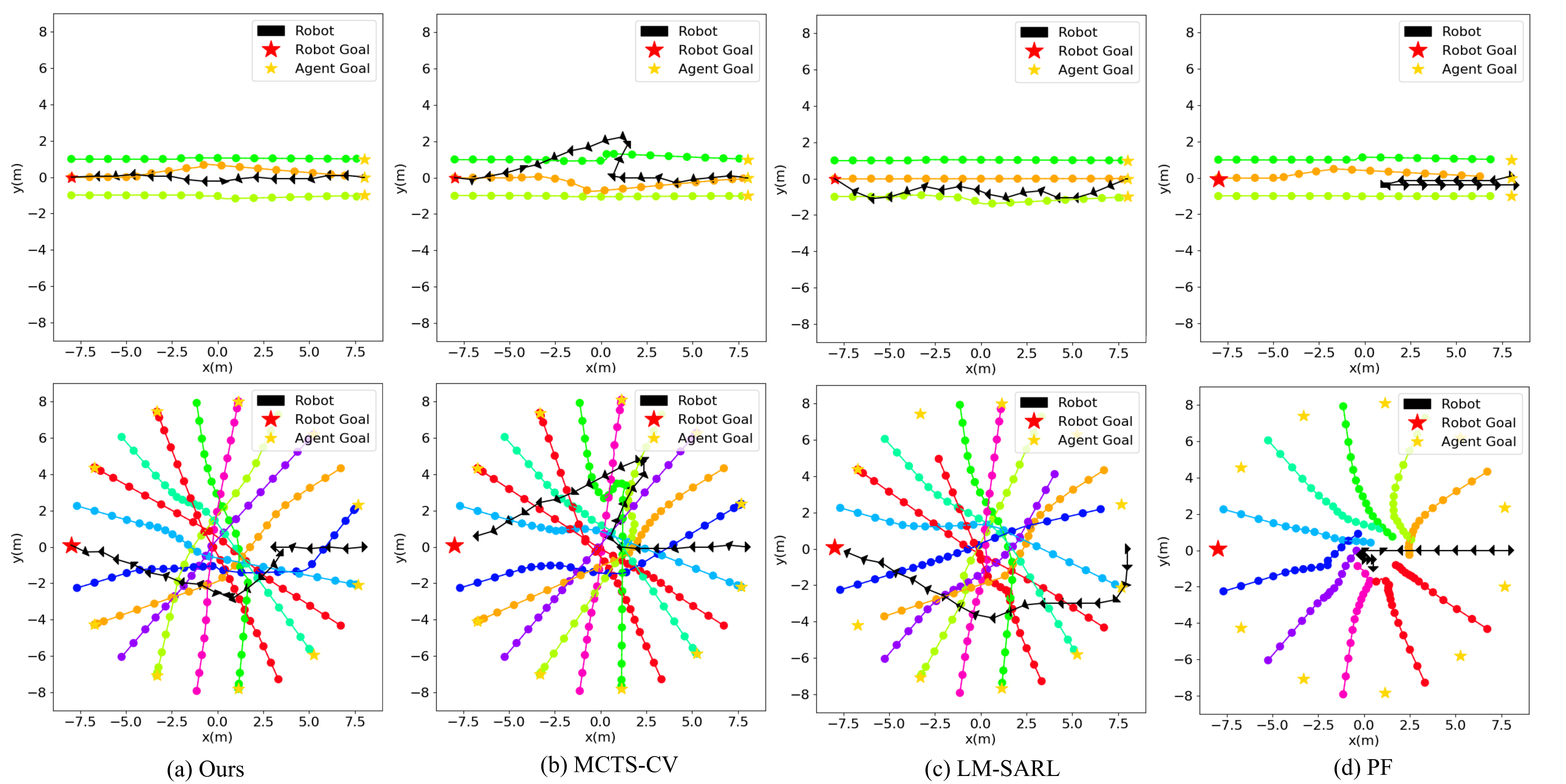}
\caption{\textit{Planner comparisons in two test cases. In the top scenario, the robot attempts to pass from right to left, through an oncoming group travelling left to right. In the bottom scenario, the robot and 10 other agents all attempt to pass to the other side of a 15m wide circle. Our method (MCTS-RNN) demonstrates ability to plan a path through the oncoming group that takes into consideration their likely response. The Potential Field method (d) displays the `frozen-robot' problem when encountering the group. All non-controlled agents are simulated using the ORCA motion model \cite{VanDenBerg2011} for both scenarios.}}
\end{figure*}
\smallskip
Planner comparisons have been performed in a simulated environment, where agents' motion is modelled using ORCA. LM-SARL has been trained as per~\cite{Chen2018}. Our planner is implemented with the following realistic assumptions.
We use a time threshold of 300ms in the MCTS based on our robot's observation frequency and discretise the action space over acceleration and yaw rate, based on real constraints of our robotic vehicle, where acc$\in$[-0.05, -0.01, 0, 0.01, 0.05] m/sec/timestep, and yaw-change$\in$[-20,-5,0,5,20] deg/timestep. For all experiments we use 50 streams for parallel Single Step Simulation.
The MCTS exploration constant $c$ has been chosen experimentally as $\sqrt{2} / 2$.

\smallskip
\textbf{Results:}
We compared performance in terms of rate of success at reaching the goal, rate of collisions with agents, average path length, computation time, and disturbance of nearby agents.
The quantitative results in \textit{Table 1} demonstrate that our planner is comparable to the state of the art methods for collision avoidance around dynamic agents. 

Disturbance was measured based on the rate that a nearby agent's absolute acceleration exceeded thresholds of 1, 0.5 and 0.25 $m/s^2$.
The use of SEF2 (\textit{Eq. 4}) allowed our approach to alter its behaviour, planning a path that minimally disturbed nearby agents without retraining. As expected, the reactive PF performs poorly, failing 50\% of the time.

Qualitative results in Fig. 4 compare the behaviour of each planner in 2 test cases. These include driving towards an oncoming group of 3 pedestrians, and a circle crossing scenario with 10 pedestrians. 
Whilst substituting constant velocity for our predictive model performed surprisingly well, both cases makes it clear that using a learnt predictive model clearly allows for better planning with consideration of agent responses.

This figure also highlights that whilst the RL method was able to find shorter paths on average, its behaviour was not always as understandable as our MCTS-RNN method, occassionally displaying oscillatory movements as shown in the top case. The `frozen-robot' problem is clearly displayed by the reactive PF planner in both cases, failing to find a path through either the oncoming group, or the circle of agents. 
\smallskip

\begin{table}[]
	\begin{tabular}{l|l|l|l|l|l|l|l|}
		\cline{2-8}
		& \multirow{4}{*}{\begin{tabular}[c]{@{}l@{}}Suc-\\ cess\\ \%\end{tabular}} & \multirow{4}{*}{\begin{tabular}[c]{@{}l@{}}Coll-\\ ision\\ \%\end{tabular}} & \multirow{4}{*}{\begin{tabular}[c]{@{}l@{}}Avg\\ Len\\ (m)\end{tabular}} & \multirow{4}{*}{\begin{tabular}[c]{@{}l@{}}Avg\\ Comp\\ Time\\ (s)\end{tabular}} & \multicolumn{3}{l|}{\multirow{3}{*}{\begin{tabular}[c]{@{}l@{}}Disturbance\\ (\% Agent Acc\\ $>$ x $m/s^2$ )\end{tabular}}} \\
		&                                                                           &                                                                             &                                                                          &                                                                                  & \multicolumn{3}{l|}{}                                                                                                              \\
		\textbf{}                             &                                                                           &                                                                             &                                                                          &                                                                                  & \multicolumn{3}{l|}{}                                                                                                              \\ \cline{1-1} \cline{6-8} 
		\multicolumn{1}{|l|}{\textbf{Method}} &                                                                           &                                                                             &                                                                          &                                                                                  & \textit{1.0}                                & \textit{0.5}                                & \textit{0.25}                                 \\ \hline
		\multicolumn{1}{|l|}{Ours}            & \textbf{98.0}                                                             & \textbf{0.0}                                                                & 20.08                                                                    & 0.3*                                                                             & 2.5                                       & 9.0                                       & 15.2                                       \\ \hline
		\multicolumn{1}{|l|}{Ours(SEF2)}      & 96.0                                                                      & \textbf{0.0}                                                                & 22.52                                                                    & 0.3*                                                                             & \textbf{1.8}                              & \textbf{5.8}                              & \textbf{11.0}                              \\ \hline
		\multicolumn{1}{|l|}{MCTS-CV}         & 93.0                                                                      & 2.6                                                                         & 20.48                                                                    & 0.3*                                                                             & 2.8                                       & 8.0                                       & 14.5                                       \\ \hline
		\multicolumn{1}{|l|}{LM-SARL}         & \textbf{98.0}                                                             & \textbf{0.0}                                                                & \textbf{19.26}                                                           & 0.20                                                                            & 2.2                                       & 9.5                                       & 15.0                                       \\ \hline
		\multicolumn{1}{|l|}{PF}              & 50.0                                                                  & 5.0                                                                         & 30.76                                                                    & \textbf{0.01}                                                                    & 2.3                                       & 11.0                                      & 16.0                                       \\ \hline
	\end{tabular}
	\caption{\textit{Planner comparisons from 500 ORCA simulations, agents numbers ranging from 2 to 12. Disturbance represents the change of agent velocity caused by the robot's motion. (*Our methods use a time budget of 300ms by design)}} \label{fig:4pics}
\end{table}




\subsection{Real-world Experiments}
We have also tested our approach in real-world experiments on a robotic platform moving through a crowd of pedestrians, instructed to walk towards chosen goals similar to the scenarios shown in \textit{Fig. 4}.
A video of these experiments is available at \href{https://youtu.be/vBPKiqtCYRU}{https://youtu.be/vBPKiqtCYRU}. Some issues were experienced during this testing, including perception issues resulting in delays in pedestrian position estimation. This led to some planning errors and near collisions when both the robot and pedestrians were travelling at high speeds.
Due to time constraints we have not performed real world comparisons of each planner included in this work, which will be required in future.

\section{Conclusion}
We have demonstrated in this work that the inclusion of a planned robot action to a predictive model can allow for more accurate prediction of agent trajectories in close range interactions.

We have also shown how this model can be used within a dynamic path planning method to achieve results comparable to state of the art. Our approach has the added benefit of being trained purely on observed data, suggesting that our model could achieve similar results in real world data, an approach not possible with reinforcement learning, in which training requires interaction with a simulated environment.

Additionally, we have demonstrated that the planner's behaviour can be altered by simply changing the state evaluation function, without requiring retraining the navigation policy, as reinforcement learning requires. This has been applied to minimising disturbance of nearby agents, and could in future be extended to any goal that aims to direct the future state of agents.

Future work will focus on extensive real world testing of our approach in comparison to existing methods, as well as improving the predictive model to account for agent-agent interactions.

\bibliographystyle{IEEEtran}
\bibliography{test}

\end{document}